\pdfoutput=1

\documentclass[11pt]{article}

\usepackage[final]{acl}

\usepackage{times}
\usepackage{latexsym}

\usepackage[T1]{fontenc}

\usepackage[utf8]{inputenc}

\usepackage{microtype}

\usepackage{inconsolata}

\usepackage[most]{tcolorbox}
\usepackage{graphicx,amssymb}
\usepackage{multirow}
\usepackage{array}
\usepackage{booktabs}
\usepackage{amsfonts}
\usepackage{bbding}
\usepackage{bm}
\usepackage{color,colortbl,xcolor}
\usepackage{stfloats}
\usepackage{enumitem}
\usepackage{nccmath}
\usepackage{subfigure}
\usepackage{soul}
\usepackage{pifont}

\definecolor{hlgreen}{HTML}{B2D5CB}
\definecolor{hlblue}{HTML}{ADD8E6}
\definecolor{bggrey}{HTML}{5E5D65}
\definecolor{bgpink}{HTML}{CEAEB9}
\definecolor{bgblue}{HTML}{8D91AA}

\title{Bridging Talk and Thought: Understanding Dialogue Dynamics Across Collaborative Problem-Solving Contexts}

\author{
Zhengyuan Liu\textsuperscript{\ding{118}*},
\ Stella Xin Yin\textsuperscript{\ding{171}}\thanks{\ Equal contribution.},
\ Min-Yen Kan\textsuperscript{\ding{169}},
\ Nancy F. Chen\textsuperscript{\ding{118}}\\
\textsuperscript{\ding{171}}Nanyang Technological University, Singapore\\
\textsuperscript{\ding{118}}Agency for Science, Technology and Research (A*STAR), Singapore\\
\textsuperscript{\ding{169}}National University of Singapore\\
\texttt{\{liu\_zhengyuan,nancy\_chen\}@a-star.edu.sg}
}

\begin{document}
\maketitle
\begin{abstract}
We present a conceptual framework for analyzing dialogue in collaborative problem-solving contexts, with an emphasis on the emerging dynamics of human-AI and multi-agent collaboration. As intelligent systems become active agents capable of autonomous reasoning and strategic cooperation, understanding the dialogic interaction during collaborative problem solving is increasingly important for optimizing and evaluating such partnerships. Our framework addresses key limitations in current analytical approaches through a hierarchical two-layer coding scheme that integrates cognitive and non-cognitive problem solving with metacognitive regulatory mechanisms. We demonstrate its effectiveness and generalizability across nine datasets spanning multiple domains, and provide insights into how humans and agents coordinate their knowledge, skills, and efforts to solve complex problems, showing in particular that metacognitive regulation can be an essential discriminator of deeper collaboration.
\end{abstract}

\section{Introduction}
Collaborative problem-solving (CPS) is a collective process where two or more agents actively engage in sharing, exchanging, and negotiating their knowledge, skills, and efforts to reach shared solutions \cite{Csapo2017}. In CPS, participants work together to achieve goals that extend beyond individual capabilities and require distributed cognition. The central component of effective collaboration is verbal interchange and communication. Through dialogue, collaborators externalize their thinking, negotiate meaning, identify problems, generate and evaluate solutions, and regulate their collective progress \cite{mercer2002words}. This dynamic interaction serves as both the vehicle and catalyst for successful collaborative efforts, highlighting the essential role of communication analysis in understanding and improving CPS. Humans engage in collaboration to overcome cognitive limitations, leverage diverse expertise, and manage complex tasks that are beyond the capabilities of any individual participant \cite{dillenbourg1999you}. The distributed nature of collaborative cognition allows groups to process more information, consider multiple viewpoints simultaneously, and generate more innovative solutions than any single person could achieve independently.

\begin{figure}[t!]
\begin{center}
\includegraphics[width=0.83\linewidth]{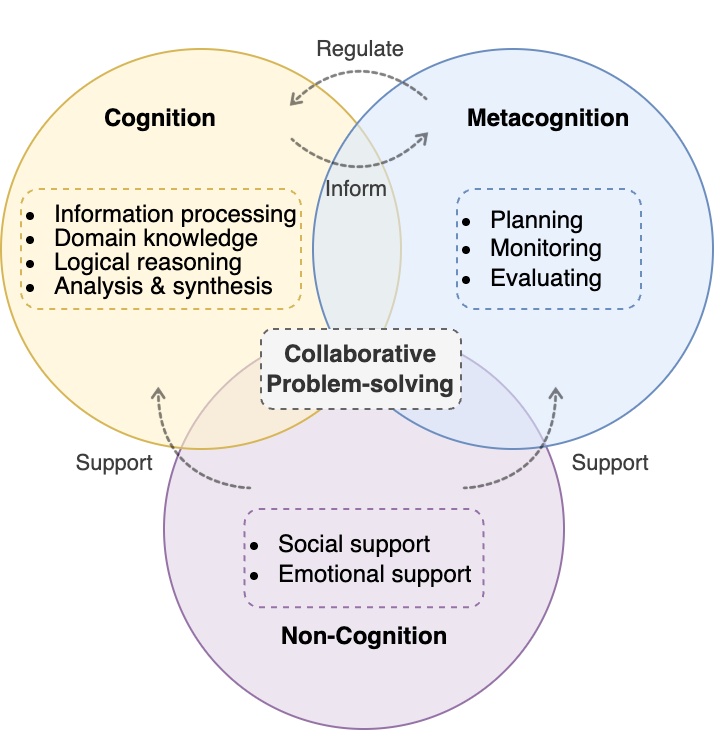}
\end{center}
\caption{A conceptual model of the dynamic interactions across cognitive, non-cognitive, and metacognitive dimensions during problem-solving.}
\label{fig_framework}
\end{figure}

Drawing on collaboration theories, successful collaborative communication relies on three interconnected components, as shown in Figure \ref{fig_framework}: \textbf{Metacognition} (regulatory behaviors that monitor and control problem-solving processes) \cite{stanton2021fostering}, \textbf{Cognition} (application of domain knowledge and reasoning) \cite{mercer2002words}, and \textbf{Non-cognition} (socio-emotional expressions that facilitate collaboration) \cite{bakhtiar2018regulation, isohätälä2020cognitive}. In this conceptual model, \textbf{Metacognition} regulates both individual and collective cognitive processes, providing oversight and control functions \cite{hadwin2017self}; \textbf{Cognition} provides the information content for metacognitive processes \cite{dillenbourg1999you}; and \textbf{Non-cognition} maintains participant engagement and manages group dynamics in order to create a supportive environment for both metacognitive and cognitive processes to function effectively \cite{naykki2014socio, bakhtiar2018regulation}.

This conceptual model is particularly critical as human--AI collaboration increasingly extends beyond brief, isolated interactions toward sustained, long-term engagements. Unlike conventional tool-use scenarios lasting only minutes, emerging collaborative tasks may span days or weeks, requiring AI agents to participate in full-pipeline workflows that involve iterative goal negotiation, continuous progress monitoring, and adaptive strategy revision across multiple sessions. In such extended contexts, the dynamic interplay of metacognitive, cognitive, and non-cognitive processes becomes essential not only for task completion but for maintaining shared understanding, sustaining motivation, and preserving joint responsibility over time \cite{hadwin2017self, dillenbourg1999you}. However, previous research reveals mixed results regarding the effectiveness of human–AI collaboration. A meta-analysis by \citet{vaccaro2024combinations} examining over 100 studies found that human--AI combinations, on average, outperformed humans working alone but did not perform better than AI working independently. Research indicates that most human--AI interactions fall into patterns of cooperation or tool use rather than genuine collaboration, such as in medical diagnosis \cite{reverberi2022experimental} and data analysis \cite{jiang2021supporting}, with AI functioning primarily as automated tools that augment human capabilities or as cooperative agents that assist with specific tasks \cite{gomez2025human}.

\begin{figure*}[t!]
\begin{center}
\includegraphics[width=0.98\linewidth]{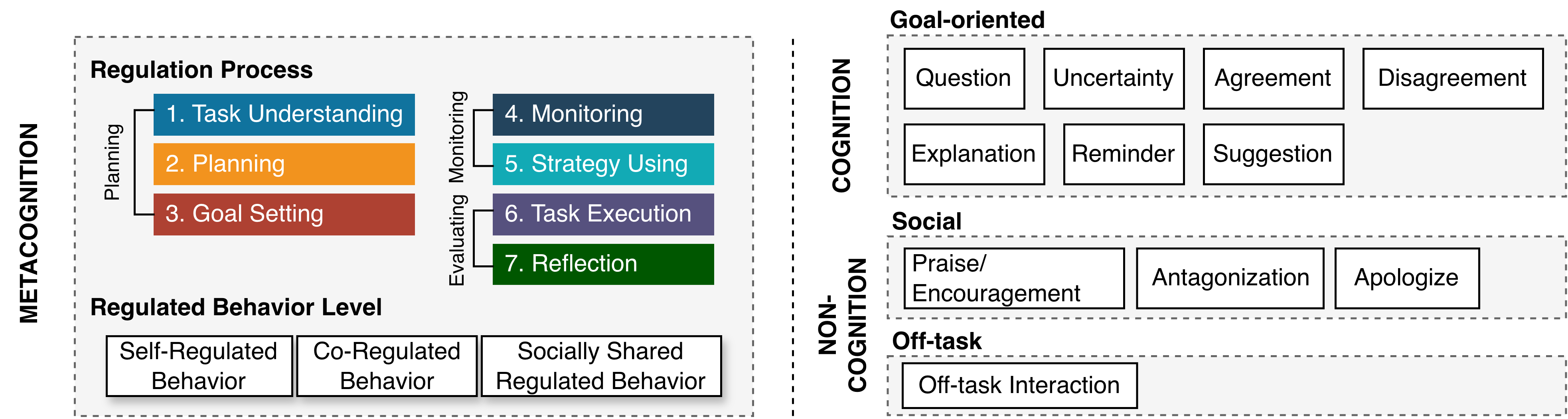}
\end{center}
\caption{A multi-structure coding scheme for collaborative dialogue analysis. The upper layer focuses on metacognitive processes and regulatory behaviors, capturing both group-level and speaker-level patterns. The lower layer examines the specific cognitive and non-cognitive behaviors at the utterance level.}
\label{fig_structure}
\end{figure*}

Since collaborative dialogue serves as the observable manifestation of collaborative processes, systematic examination of communication patterns could help to identify whether certain interactions contribute to mutual engagement toward shared goals, complementary contributions of unique expertise, adaptive responsiveness to partner actions, negotiated shared understanding, and joint responsibility for outcomes \cite{galescu-etal-2018-cogent,cohen-2019-foundations,liu-chen-2021-improving,yin2025scaling}. Conversely, asymmetric dialogue patterns, where one party drives goal-setting, monitoring, and evaluation while the other responds reactively, indicate cooperation or tool use rather than true collaboration. Such understanding can help distinguish and measure the degree of collaboration in human--AI contexts, explain why human--AI collaboration succeeds or fails, and inform the design of systems that achieve genuine collaborative partnerships.

Therefore, in this work, we introduce a multi-structure coding scheme for collaborative dialogue analysis (Figure \ref{fig_structure}) that characterizes collaborative problem-solving across \textbf{Metacognition}, \textbf{Cognition}, and \textbf{Non-cognition} dimensions. The coding scheme classifies the dialogue patterns at the utterance level and assesses each member's contribution and distribution during collaboration activities. This approach differentiates collaborative interactions from mere cooperative exchanges or tool use scenarios by identifying when participants demonstrate asymmetry across the three dimensions. We conducted experiments on nine CPS datasets in multiple scenarios. We observed that structured environments often induce functional asymmetry where one participant assumes a metacognitive leadership role, indicating that collaborative patterns are context-dependent and shaped by task structure. Furthermore, our analysis of human-AI interactions shows that the imbalance remains in common collaborative settings: human users dominate self-regulation, whereas AI systems are incline to reactive and co-regulated behaviors, which helps explain why human-AI collaboration often fails to achieve genuine partnership and clarifies the current role of AI as a responsive assistant. The framework and coding scheme can be generalized to different domains, particularly, validating metacognitive regulation as an essential discriminator for accurately modeling collaboration. This work also offer practical guidance for developing more synergetic human--AI collaborative systems, such as detecting regulatory imbalance in real time, triggering adaptive scaffolding to balance contributions, and guiding the design of AI agents that can initiate and share metacognitive regulation rather than merely execute assigned tasks.

\section{Related Work}

Natural language processing research has long focused on modeling dialogue structures and discourse acts to understand communicative intents \cite{asher-etal-2016-discourse,budzianowski-etal-2018-multiwoz}. From statistical to neural models and large language models, various methods are proposed to classify conversational intents \cite{williams-etal-2013-dialog}, track dialogue states \cite{mrksic-etal-2017-neural,aksu-etal-2022-nshot}, analyze discourse \cite{asher-etal-2016-discourse,liu-chen-2021-improving}, and evaluate interaction quality in complex, task-oriented contexts \cite{yin2025scaling}.
In linguistics and the learning sciences, collaborative patterns are traditionally analyzed through the lens of exploratory talk and socio-emotional regulation. Foundational studies highlight how shared understanding is dynamically negotiated via epistemic interactions, linguistic alignment, and metacognitive scaffolding \cite{mercer2002words,hadwin2011self,dillenbourg1999you}.
As systems transition from passive tools to active conversational agents, research increasingly evaluates interaction symmetry and joint problem-solving dynamics \cite{galescu-etal-2018-cogent,gomez2025human,yin2025scaling,tan-etal-2025-persuasion}. Recent works formalize decision-oriented dialogues where humans and AI must combine disparate information to negotiate shared goals \cite{lin2024decision}. However, meta-analyses reveal that human--AI pairs often fail to outperform the best individual agent, frequently devolving into asymmetric tool-use rather than real partnership \cite{jiang2021supporting, reverberi2022experimental, vaccaro2024combinations}.
To close this gap, recent work has turned to proactive, metacognition-supporting agents \cite{gmeiner2025exploring, mukhopadhyay2026exploring} and multi-agent collaboration benchmarks \cite{zhu2025multiagentbench}, seeking to move interaction from cognitive offloading toward shared initiative.
In this work, we bridge these computational and cognitive perspectives with a unified coding scheme to capture multi-dimensional interactions underlying collaborative problem-solving.

\section{CPS Conceptual Model and Coding Scheme}
Collaborative problem solving (CPS) is a collective process where individuals work together to achieve a shared goal. In CPS, two or more agents actively engage in sharing, exchanging, and negotiating their knowledge, skills, and efforts for solutions \cite{Csapo2017}. The communication nature of group-based work highlights cognitive, social, and metacognitive dimensions that dynamically interact during the problem-solving process \cite{dillenbourg1999you, nelson2013collaborative, woolley2010evidence}. We introduce a multi-structure coding scheme (Figure \ref{fig_structure}) to systematically track the above three dimensions: metacognitive awareness through planning, monitoring, and evaluating regulation processes at three behavioral levels; cognitive processes through seven types of goal-oriented interactions; and non-cognitive elements through three types of social interactions. The following subsections explain the theoretical foundations and key components of each dimension in detail.

\begin{table*}[t]
\centering
\small
\resizebox{1.0\linewidth}{!}
{
\begin{tabular}{p{2cm}p{3cm}p{10cm}}
\toprule
\textbf{Metacognition} & \textbf{Regulation Process} & \textbf{Definition and Examples Corresponding to the 8-Stage CPS Process} \\
\midrule
Planning & \textcircled{1} Task Understanding & Identifying and defining the problem (Stage I: ``What's going on?''); Understanding what information is already available about the problem (Stage II: ``What do we know?''); Investigating root causes of the problem (Stage III: ``What are the root causes?'') \\
 & \textcircled{2} Planning & Establishing team structure and distributing tasks strategically (Stage IV: ``What could we do?'') \\
 & \textcircled{3} Goal Setting & Setting objectives for problem resolution; Determining criteria for selecting the best solution (Stage V: ``What's the best thing to do?'') \\
\midrule
Monitoring & \textcircled{4} Monitoring & Tracking progress, summarizing completed work, and identifying remaining tasks \\
 & \textcircled{5} Strategy Use & Proposing potential solutions; Evaluating options and selecting the best solutions (Stage VI: ``How do we go about it?'') \\
\midrule
Evaluating & \textcircled{6} Task Execution & Implementing selected solutions (Stage VII: ``Have we solved the problem?'') \\
 & \textcircled{7} Reflection & Reflecting and making adjustments for continuous improvement (Stage VIII: ``Can we improve on what we have done?'') \\
\bottomrule
\end{tabular}
}
\caption{Seven regulation processes in collaborative problem-solving contexts.}
\label{tab:Meta_framework}
\end{table*}

\subsection{Metacognition}
Metacognition refers to the process by which participants monitor and regulate their own thinking and behaviors. In collaborative problem-solving contexts, individuals employ knowledge of the task and problem-solving strategies to plan their work, monitor their progress toward goals, and evaluate outcomes \cite{national2000inquiry, berliner1994expertise}. We examine metacognition through two complementary lenses: \S\ref{sec_2.1.1} regulation processes (planning, monitoring, and evaluating) and \S\ref{sec_2.1.2} behavioral levels that distinguish how regulation is initiated and maintained (self-regulated, co-regulated, and socially shared-regulated behaviors).

\begin{figure*}[t!]
\begin{center}
\includegraphics[width=1.0\linewidth]{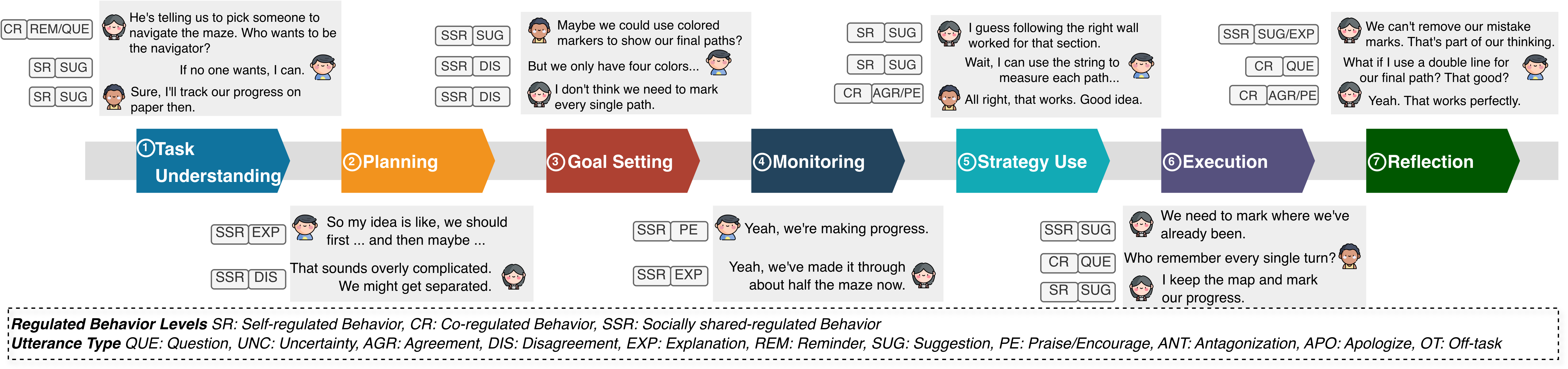}
\end{center}
\caption{A dialogue example of three individuals collaboratively solving a maze puzzle. The conversation is segmented into seven sequential regulation processes from task understanding to reflection, with each utterance categorized according to regulated behavior levels and cognitive/non-cognitive utterance types.}
\label{fig_example}
\end{figure*}

\subsubsection{Regulation Process} \label{sec_2.1.1}
Metacognition typically involves planning, monitoring, and evaluating \cite{stanton2021fostering}. Building on \citeauthor{Winne1998}'s \citeyearpar{Winne1998} metacognitive model, we decompose it into seven sequential processes (see Table \ref{tab:Meta_framework}). To analyze specific interaction patterns and behaviors within each process, we integrated this framework with the eight-stage CPS process model \cite{nelson2013collaborative}, developing a coding scheme that examines how groups collectively solve problems through dialogue, from initial problem identification to solution evaluation and refinement. Here, we show a dialogue example (Figure \ref{fig_example}) to illustrate how these seven regulation processes unfold in collaborative dialogue. In this example, three students work together to solve a maze puzzle task and identify the shortest route. Their conversation reveals how the team moves through each regulation phase, from initial task comprehension to final reflection on their collaborative process.

\noindent\textbf{Planning} \textcircled{1}Task Understanding includes identifying the problem, assessing available information, and investigating root causes. In our maze puzzle example, this process begins when group members identify problems, as the female student inquires, ``\textit{He's telling us to pick someone to navigate the maze. Who wants to be the navigator?}'' \textcircled{2}Planning then builds on this understanding by distributing tasks strategically. One team member suggests, ``\textit{So my idea is like, we should first check if there are multiple entrances to the maze, and then maybe split up to try different paths. If we hit a dead end, we can mark it on our map and backtrack...}'' \textcircled{3}Goal Setting completes the planning process   by establishing specific objectives and criteria for optimal solutions, ``\textit{Maybe we could use colored markers to show our final paths? And demonstrate it as the shortest one?}''

\noindent\textbf{Monitoring} \textcircled{4}Monitoring refers to tracking progress and identifying remaining tasks. In our example, we observe one member tracks the progress and reminds other members that ``\textit{Yeah, we've made it through about half the maze now.}'' \textcircled{5}Strategy Use focuses on proposing solutions from different perspectives, such as: ``\textit{I guess following the right wall worked for that section.}''

\noindent\textbf{Evaluating} \textcircled{6}Task Execution refers to implementing selected solutions, as one participant comments, ``\textit{We need to mark where we've already been.}'' After that, \textcircled{7}Reflection involves group members assessing outcomes and reflecting on the process for further improvements. For example, ``\textit{We should probably mark our final route more clearly. Like this. So that we don't confuse it with the dead ends we explored.}''

It is important to note that these seven regulation processes do not operate as a rigid, linear sequence in practice. In complex problem-solving contexts, groups frequently revisit, skip, or cycle through stages in response to emerging challenges and new information. For instance, a team may return from \textcircled{6}Task Execution back to \textcircled{5}Strategy Use when an implemented solution proves ineffective, or loop back to \textcircled{1}Task Understanding mid-way through \textcircled{4}Monitoring when the group realizes the problem has been misidentified. In our maze puzzle example, this non-linearity is evident when a member's reflection simultaneously triggers a re-evaluation of their strategy and a revision of their shared goal.

\subsubsection{Regulated Behavior Levels} \label{sec_2.1.2}
The regulation processes described above focus on \textit{what} problem-solving steps teams follow, while the regulated behavior levels examine \textit{how} group members actually coordinate and work together during collaboration. From speaker perspective, these regulatory behaviors occur at three distinct levels (individual, interpersonal, and collective), depending on who takes the lead in guiding the group's efforts. Unlike the sequential problem-solving processes that track the team's progress through tasks, these behavior levels reveal the social dynamics of how participants share responsibility, influence each other, and collectively manage their collaborative work \cite{hadwin2017self} (see Table \ref{tab:regulation_examples}).

At the individual level, \textbf{Self-regulated Behavior} (SR) is often characterized as an active and goal-directed process in which individuals are portrayed as active intentional regulators of their own cognition, metacognition, motivation, and behavior \cite{pintrich2000multiple}. In CPS contexts, this manifests when individuals assess their own knowledge gaps, recognize their cognitive limitations, or take personal responsibility for specific task components. Linguistically, self-regulation is marked by first-person singular expressions, with an emphasis on ``I'' statements (e.g., ``\textit{I need to do this},'' ``\textit{I don't understand this concept.}''). In our maze example, ``\textit{I'll track our progress on paper then}'' demonstrates personal responsibility and active metacognitive planning about how individual actions support group goals.

At the interpersonal level, \textbf{Co-regulated Behavior} (CR) involves the scaffolding of regulatory processes between group members, where one's regulatory activity supports or constrains another's learning and performance \cite{hadwin2011self}. In the CPS context, co-regulation serves critical functions in collaboration because it distributes cognitive load, provides external monitoring when self-regulation fails, and creates opportunities for regulatory skill transfer between participants \cite{miller2015scripting}. Beyond the linguistic markers of second-person address (``\textit{Do you think...?}'' or ``\textit{Can you...?}''), co-regulation involves sophisticated social cognition. For example, in the maze puzzle task, interpersonal regulation happens when one member asks, ``\textit{Who remembers every single turn?}'' prompting a specific contribution from others, and when another responds to a suggestion with, ``\textit{All right, that works. Good idea.}'' These instances demonstrate group members are strategically directing cognitive resources and compensating for potential memory limitations in a group.

At the collective level, \textbf{Socially shared-regulated Behavior} (SSR) represents the most integrated form of regulation, emerging when group members collaboratively negotiate shared perceptions and goals. This collective regulation involves coordinated strategic planning, joint monitoring of group progress, and collaborative adaptation when needed to optimize performance \cite{hadwin2011self}. Unlike self-regulation and co-regulation, socially shared regulation distributes regulatory responsibility across the entire group, creating a truly collective ownership of the problem-solving process. Linguistically, this form of regulation is characterized by inclusive first-person plural language (``we,'' ``us,'' ``our'') and collaborative decision-making expressions. For example, ``\textit{What if we set a time limit for each section of the discussion?}'' Our maze example includes numerous instances of shared regulation behaviors, such as ``\textit{We might get separated.}'' and ``\textit{I don't think we need to mark every single path.}'' These exchanges show how the group constructs understanding and coordinates its regulatory activities toward shared goals.

We emphasize that all three regulated behavior levels are central to effective CPS. Empirical evidence shows that over-reliance on self-regulation leads to coordination failures, while excessive co-regulation creates dependency relationships that limit group autonomy \cite{miller2015scripting}. Conversely, groups that successfully integrate all three levels demonstrate better problem-solving performance and more sustainable collaborative relationships \cite{isohätälä2020cognitive}. We believe that effective CPS requires dynamic interplay among regulatory processes, where each participant must simultaneously maintain their own self-regulation, provide regulatory guidance to others through co-regulation, and participate in collective regulatory efforts through socially shared regulation \cite{hadwin2017self}.

\begin{table*}[t!]
\centering
\small
\resizebox{1.0\linewidth}{!}
{
\begin{tabular}{p{2.1cm}p{3.3cm}p{10cm}}
\toprule
\textbf{Interaction} & \textbf{Utterance Type} & \textbf{Definition} \\
\midrule
Goal-oriented & 1. Question (QUE) & Any inquiry or expression of concern seeking clarification, confirmation, or additional information about the task. \\
 & 2. Uncertainty (UNC) & Expression of doubt, hesitation, or lack of confidence about ideas, decisions, or information, often indicating a need for guidance or support from others. \\
 & 3. Agreement (AGR) & Expression of consensus, acceptance, or alignment with another's opinion, suggestion, or idea. \\
 & 4. Disagreement (DIS) & Expression of dissent, rejection, or misalignment with another's opinion, suggesting an alternative perspective. \\
 & 5. Explanation (EXP) & Providing detailed clarification, elaboration, or reasoning to enhance understanding of concepts or decisions. \\
 & 6. Reminder (REM) & Redirecting attention to task requirements, time constraints, or previously agreed-upon goals to maintain focus. \\
 & 7. Suggestion (SUG) & Proposing new ideas, approaches, or solutions related to the task. \\
\midrule
Social & 8. Praise/Encourage (PE) & Offering positive feedback, validation, or motivation to boost team morale and confidence. \\
 & 9. Antagonization (ANT) & Making harmful comments, instigating conflict, criticizing personal attributes, or showing irritation toward collaborators. \\
 & 10. Apologize (APO) & Acknowledging mistakes, misunderstandings, or inappropriate behaviors with an expression of regret. \\
\midrule
Off-task & 11. Off-task (OT) & Any utterance unrelated to the current task or collaboration objectives. \\
\bottomrule
\end{tabular}
}
\caption{Cognitive/Non-cognitive utterance types in collaborative problem-solving contexts.}
\label{tab:Cog_table}
\end{table*}

\subsection{Cognition}
Cognition forms the foundation of the actual thinking processes that team members use to understand problems, generate solutions, and make decisions (Figure \ref{fig_framework}). This includes analyzing information, applying domain knowledge, and logical reasoning. To understand group members' cognitive processes from utterance level, we build on \citeauthor{mercer2002words} \citeyearpar{mercer2002words} exploratory talk and previous empirical studies to classify the goal-oriented utterances into seven types (see Table \ref{tab:Cog_table} for coding scheme details). They are question (QUE), uncertainty (UNC), agreement (AGR), disagreement (DIS), explanation (EXP), reminder (REM), and suggestion (SUG). These utterance types represent how group members critically and constructively build upon each other's contributions and create a progressive dialogue that leads to enhanced reasoning and deeper conceptual understanding \cite{mercer2002words}. We apply this coding scheme and label each utterance in our maze puzzle example (Figure \ref{fig_example}).

\subsection{Non-cognition}
According to Bakhtiar et al. \citeyearpar{bakhtiar2018regulation}, social interactions are purposeful interchanges among group members that shape perceptions of emotions and socio-emotional climate within the group. Positive interaction, such as encouragement or positive feedback on group work, has been shown to facilitate productive collaboration \cite{isohätälä2020cognitive}. In contrast, negative interaction, such as discouraging other members from participating or disrespecting other group members, can have adverse consequences on group cohesion, commitment, satisfaction, and performance \cite{naykki2014socio}. Although social interactions are not directly related to problem-solving tasks, they are essential requirements for supporting content-related discussion and group cohesion. In CPS, we define three types of utterances: praise and encouragement (PE), antagonistic comments (ANT), and apologies (APO) (see Table \ref{tab:Cog_table} for coding scheme details).

\begin{figure*}[t!]
\begin{center}
\includegraphics[width=1.0\linewidth]{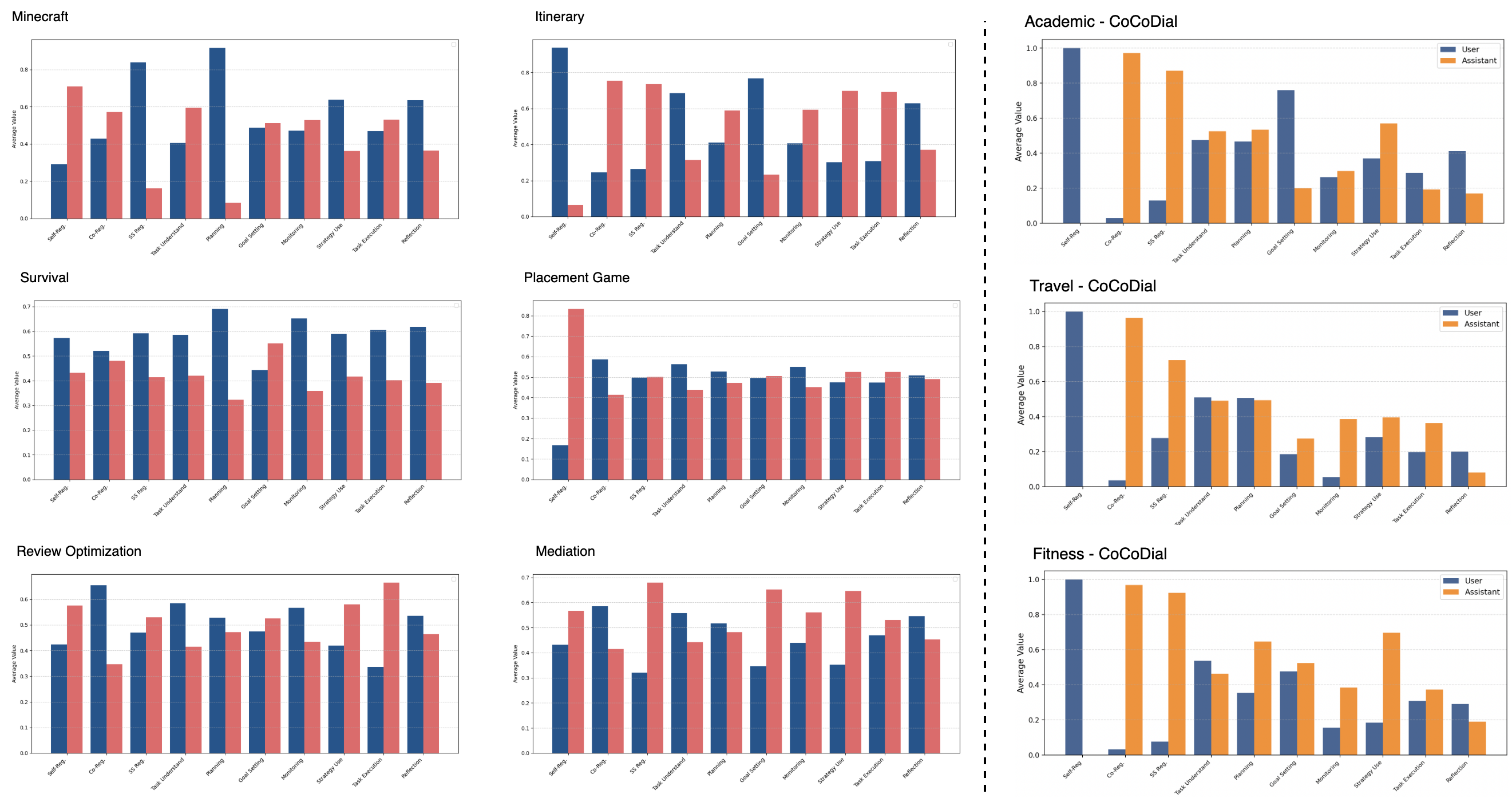}
\end{center}
\vspace{-0.2cm}
\caption{Regulation process characterization in different data: human-human and human-AI. The X axis denotes (from left to right): three regulated behavior levels (\textit{Self-, Co-regulated, and Socially Shared Regulated}), and seven metacognitive regulation processes (\textit{Task Understanding, Planning, Goal Setting, Monitoring, Strategy Use, Task Execution, and Reflection}).}
\label{fig_bar_chart}
\vspace{-0.2cm}
\end{figure*}

In addition to the above dimensions, we also highlight off-task behavior as an indispensable part yet often overlooked component of collaborative dialogue. Off-task interaction refers to utterances unrelated to the current task or objectives, which provide important context for understanding engagement patterns and potential distractions. While off-task interaction may impede problem solving, it could have positive roles in CPS if it supports the socio-affective dimensions of interaction (e.g., ``\textit{Anyone else surviving on coffee alone today?}'' or ``\textit{That reminds me of that maze in The Shining—hopefully ours is less terrifying!}''). This argument is supported by previous research \cite{langer2020exploring} suggesting that there is no clear distinction between ‘off’ and ‘on’ task talk and that group members might weave between ‘on’ and ‘off’ task conversation while remaining engaged with tasks. Such interweaving of interaction contributes to a more productive CPS environment. More importantly, accurately distinguishing between positive and negative off-task behavior cannot be reliably achieved by examining a single utterance in isolation. Therefore, instead of relying solely on the utterance itself, our coding scheme incorporates the surrounding conversational context that considers both the preceding and following utterances to capture the functional role of each off-task turn within the broader dialogue.

\begin{figure*}[t!]
\begin{center}
\includegraphics[width=1.0\linewidth]{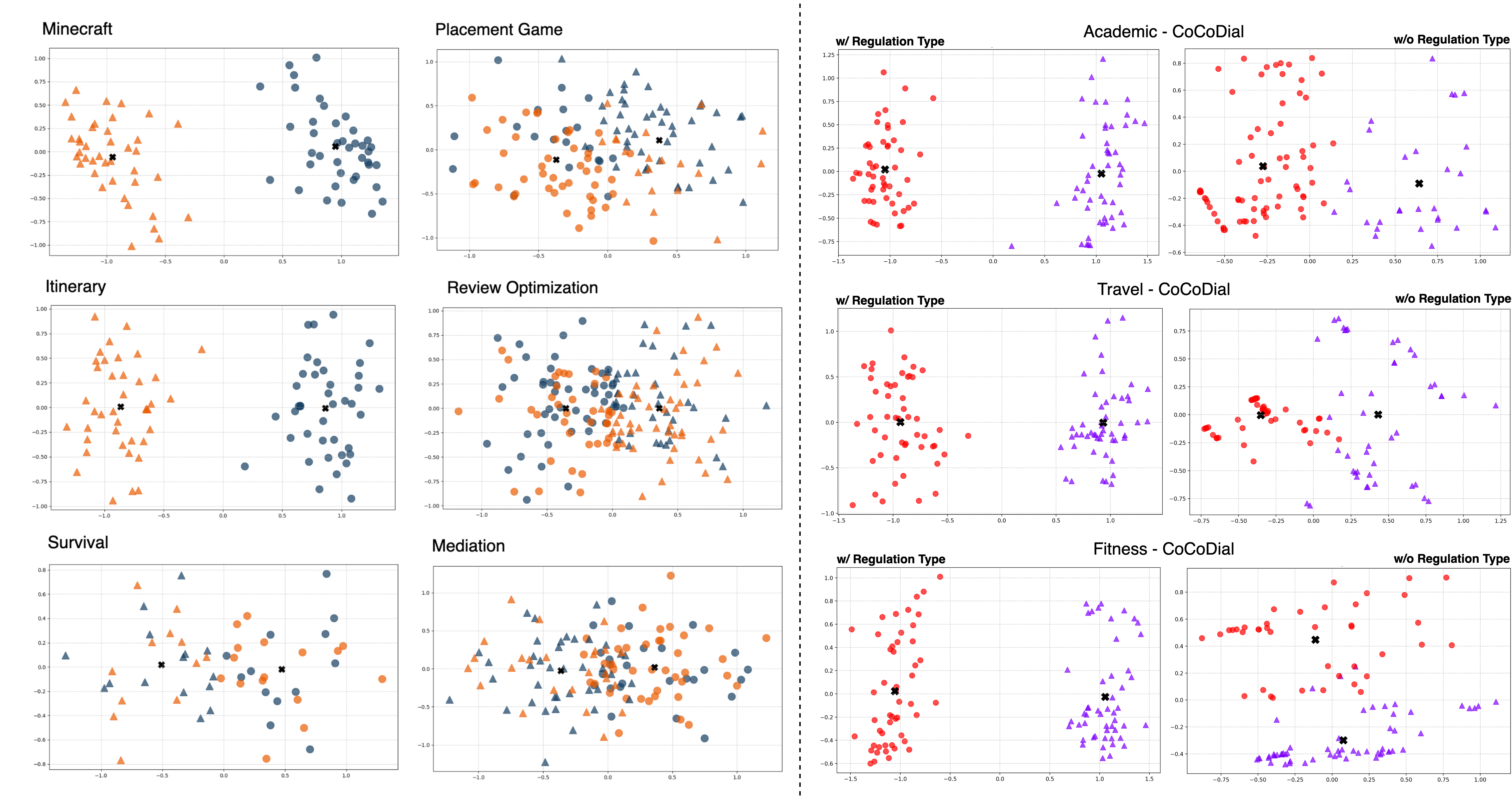}
\end{center}
\vspace{-0.2cm}
\caption{Clustering analysis of collaborative roles in different datasets based on categorization across metacognitive, cognitive, and non-cognitive dimensions. Colors of markers represent explicit roles, while the shapes (e.g., triangle and circle) represent embedded roles within the collaborative dialogue. The clustering patterns reveal how different combinations of communicative elements correspond to distinct collaborative functions.}
\label{fig_speaker_cluster}
\end{figure*}

\section{Collaborative Dialogue Analysis}

\subsection{Cross-domain Characteristics}
Collaborative problem solving manifests distinctively across various contexts, each with unique characteristics that shape interaction patterns and information processing behaviors. As shown in Table \ref{tab:cps-contexts}, educational collaboration, focusing on knowledge construction and skill development, typically features structured communication within extended timeframes, where process learning is often as valued as outcomes \cite{roschelle1995construction}. In workplace settings, CPS is oriented toward organizational goals and problem resolution, characterized by both formal and informal communication channels with strong results orientation \cite{kauffeld2012meetings}. Crisis response contexts involve time-critical, high-stakes situations requiring protocol-driven communication and immediate problem resolution \cite{kapucu2011collaborative}. Game-based collaboration offers another distinctive environment where problem solving is constrained by game rules, often time-limited, with dual focus on process enjoyment and winning \cite{zagal2006collaborative}. 
Generally, the \textbf{explicit roles} refer to formally assigned positions in collaborative tasks (e.g., designated leader, navigator in the maze example), while \textbf{embedded roles} emerge naturally through dialogue patterns without formal assignment (e.g., someone consistently monitoring progress or providing encouragement).

\subsection{Corpus Selection}
We select 6 human-human and 3 human-AI datasets based on our CPS framework:
(\textbf{D1}) The GAP Corpus \cite{gapcorpus18} is based on the Winter Survival Task, where people rank 15 items by importance in a hypothetical emergent scenario.
(\textbf{D2}) Game-based collaboration dataset \cite{jeknic-etal-2024-dialogue} contains crowd-sourced interactions between pairs playing a 2D object placement game specifically designed to elicit balanced collaboration. 
(\textbf{D3-D5}) Three datasets from \citet{lin2024decision} include three decision-making tasks of workplace collaboration: paper reviewer assignment, multi-step itinerary planning, and group travel mediation.\footnote{In the mediation scenario, there are three speakers while two speakers play the same role. Therefore, we select the two distinct roles to demonstrate their collaborative characteristics.} (\textbf{D6}) MSDC \cite{narayan2019collaborative}, which provides discourse-annotated collaborative interactions within Minecraft.
(\textbf{D7-D9}) Progressive human-AI cognitive alignment in exploratory and personalized scenarios (i.e., academic, travel, and fitness planning).\footnote{https://github.com/arenceLi/CoCoDial}
These datasets were selected based on three criteria: 1) they contain problem-solving interactions with clearly defined goals, 2) they represent diverse contexts, and 3) they include sufficient dialogue data to enable meaningful analysis of collaborative dynamics.

\subsection{Dialogue Annotation}
For each dataset, we selected 20 representative samples for human annotation. Three trained annotators independently coded each utterance for metacognitive regulation types (Table \ref{tab:Meta_framework}), regulated behavior level (SR, CR, and SSR), and utterance types (Table \ref{tab:Cog_table}). Inter-rater reliability was calculated using Cohen's kappa, with disagreement resolved through consensus discussion. Final kappa values ranged from 0.72 to 0.85 across coding dimensions, indicating reasonable agreement.

\subsection{LLM-based Utterance Analysis}
To efficiently analyze dialogue at scale across multiple datasets, we used GPT-4o for LLM-as-a-judge \cite{li-etal-2023-coannotating}.\footnote{https://openai.com/index/gpt-4o-system-card/} We created detailed prompts that incorporated our annotation guideline with explicit descriptions and examples of each category (see examples in Table \ref{appendix-inst-example}). For each utterance in context, the model was instructed to classify: 1) which regulation process the utterance belonged to, 2) which regulated behavior level it represented, and 3) which specific utterance type it exemplified. To ensure reliability, we validated this approach by comparing model classifications against human annotations on a held-out subset of dialogues, achieving substantial agreement with Cohen's kappa values ranging from 0.69 to 0.75 across the different coding dimensions.

\subsection{Regulation Process Characterization}
We transformed the utterance-level annotations into speaker-level vectors to capture their collaborative patterns. For each dataset, we first calculate speakers' utterance ratio of each dialogue based on their regulated behavior labels and regulation process labels. We then take the averaged numbers across all dialogues to demonstrate their holistic pattern on the specific CPS task.

Figure \ref{fig_bar_chart} presents results on the 9 human-human and human-AI datasets. We found that when participants contribute equally to collaboration, their dialogue exhibits balanced distributions across metacognitive, cognitive, and social dimensions (e.g., Review Optimization and Mediation). In contrast, collaborative interaction in Minecraft and Itinerary shows imbalance in both regulation behavior and processes. This asymmetry extends across multiple metacognitive processes, particularly in Planning, Goal Setting, and Task Execution, suggesting that one collaborator assumes a stronger directive and organizational role while the other adopts a more responsive and executory function. We also observe that in the Placement Game, asymmetry appears when one participant demonstrates significantly higher self-regulated behaviors, while maintaining balanced patterns across other patterns. This sharp contrast suggests that leadership or dominant role in metacognitive regulation can coexist with equitable cognitive contributions.
Moreover, on the CoCoDial datasets, we observed that human-AI interactions show a very imbalanced distribution across the three regulated behavior levels, with humans dominating self-regulation while assistants kept Co- and Socially Shared Regulated. This reflects the AI's inherently reactive nature, where responses are always prompted by human input rather than autonomous initiative.
When AI seldom initiates planning, goal-setting, or monitoring, the human bears the full regulatory burden \cite{fan2025beware}. Detecting this asymmetry is a prerequisite for pushing systems to share regulation rather than passively await instruction \cite{gmeiner2025exploring}.

These findings have several implications for future collaborative dialogue analysis. The variation in regulatory balance across contexts suggests that effective CPS manifests differently depending on task structure and role configurations. The identification of asymmetric regulation patterns also opens possibilities for real-time collaboration assessment, where shifts in regulated behavior levels could trigger adaptive scaffolding to rebalance contributions before unproductive collaboration occurs.

\subsection{Collaborative Role Clustering}
To further study how consistent the speakers' collaborative role is in each CPS task, we performed a clustering analysis on the dialogue-level collaborative pattern. We extract regulation and interaction types from speaker utterances, calculate and vectorize cross-speaker distributions, and then apply K-means to cluster their roles. As shown in Figure \ref{fig_speaker_cluster}, the separation of collaborative roles varies significantly based on task structure and the inclusion of metacognitive regulation features.
Human-human interactions in the left part demonstrate that highly structured tasks, such as Minecraft and Itinerary, indicate a clear division between speakers. In contrast, tasks requiring more negotiation, like Mediation or the Placement Game, showing overlapped clusters and more symmetrical roles.
Moreover, the CoCoDial ablation study on regulation-type labeling demonstrates the critical impact of metacognitive regulation. Including regulation types produces tightly grouped, distinct clusters, while removing them results in highly mixed, diffuse distributions. This confirms that regulation behaviors are essential discriminators for more accurately defining and modeling collaboration.

\section{Conclusion}
In this paper, we introduced a comprehensive analytical framework that conceptualizes collaborative problem solving as the dynamic interplay of metacognition, cognition, and non-cognition dimensions. Our analysis across diverse contexts reveals that effective collaboration features balanced distributions of metacognitive regulation processes and cognitive contributions. The hierarchical structure of our framework provides a principled approach for evaluating human-human and human--AI partnerships, and points toward a future where both human and AI agents contribute their unique capabilities while maintaining shared agency and responsibility, advancing both theoretical foundations and practical development of effective collaborative systems.

\section*{Limitations}
All samples used and generated in this work are in English; thus, to apply the model to other languages, it will require additional data pre-processing steps on the specified language or using multilingual language backbones.
We are aware that it remains an open problem to mitigate hallucinations and biases in large language models, which may cause communication issues in human-machine interaction. Of course, current models and laboratory experiments are always limited in this or similar ways. We do not foresee any unethical uses of our proposed methods or their underlying tools, but hope that they will contribute to reducing incorrect system outputs.

\section*{Ethics and Impact Statement}
We acknowledge that all of the co-authors of this work are aware of the provided ACL Code of Ethics and honor the code of conduct. In our experiments, models are applied under proper license. All data used in this work are only for academic research purposes and should not be used outside of academic research contexts. Our proposed methodology in general does not create a direct societal consequence and are intended to be used to improve the performance, robustness, and safety of AI systems.

\section*{Acknowledgements}
This research is supported by the Agency for Science, Technology and Research (A*STAR), Singapore, and the National Research Foundation Singapore under the AI Singapore Programme (AISG Award No: AISG3-RPGV-2025-018). We thank the anonymous reviewers for their precious feedback to help improve and extend this piece of work.

\bibliography{custom}

\clearpage

\appendix

\begin{table*}[t!]
\centering
\small
\begin{tabular}{p{2.5cm}p{3.5cm}p{3.5cm}p{4.5cm}}
\toprule
\textbf{Regulation Types} & \textbf{Self-Regulated Level} & \textbf{Co-Regulated Level} & \textbf{Socially Shared Regulated Level} \\
\midrule
Task Understanding & ``I need to figure out what this problem is asking for.'' & ``Do you agree that we need to understand what's being asked?'' & ``I think we should also check what resources are already available to us.'' \\
\midrule
Planning & ``I'm going to look at the data first.'' & ``Let's split up the work. Maria, could you handle the research part?'' & ``We could create a shared document to track progress. Does that work?'' \\
\midrule
Goal Setting & ``I will find a solution that reduces costs by at least 15\%.'' & ``Our objective should be to reduce costs. Does everyone agree we should aim for a 15\% reduction?'' & ``I think our primary goal should be cost reduction, but we should also consider customer satisfaction.'' \\
\midrule
Monitoring & ``I've completed the first part of the analysis.'' & ``Have you finished the analysis section, Sam?'' & ``I noticed we still need to address the risk section. Should we reassign that or adjust our timeline?'' \\
\midrule
Strategy Use & ``I think approach A is the best option because it addresses our key constraints.'' & ``We could try solution A or B. Which do you all prefer? I think A might work better for our needs.'' & ``A is more cost-effective but B offers better scalability. What if we combine elements from both? Maria, you mentioned a hybrid approach earlier - could we integrate that with these options?'' \\
\midrule
Task Execution & ``I calculated the data based on our plan.'' & ``Can you prepare the slides for reporting our findings?'' & ``How about we create slides on Google Drive? We might need to adjust our approach based on what we discover during implementation.'' \\
\midrule
Reflection & ``I think I can recreate the scatter plot to visualize the result.'' & ``Our solution didn't work as expected. Should we try the alternative method we discussed earlier?'' & ``I think our solution didn't fully address the problem. Probably because we overlooked the seasonal variations. I remember Sarah mentioned this point before, right?'' \\
\bottomrule
\end{tabular}
\caption{Examples of different regulation levels in CPS}
\label{tab:regulation_examples}
\end{table*}


\begin{table*}[t!]
\centering
\small
\resizebox{1.0\linewidth}{!}
{
\begin{tabular}{p{2cm}p{4cm}p{4cm}p{2.5cm}p{2.5cm}}
\toprule
\textbf{CPS Context} & \textbf{Definition} & \textbf{Scenario} & \textbf{Role Definition} & \textbf{Outcome Focus} \\
\midrule

\textbf{Educational Collaboration} & 
CPS in formal/informal learning environments for knowledge construction \& skill development & 
Knowledge construction focused; scaffolding; emphasis on reflection and metacognition & 
Teacher assigned or rotated roles & 
Learning as important as outcome \\
\midrule

\textbf{Workplace Collaboration} & 
CPS in professional settings for organizational goals, innovation, and problem resolution & 
Efficiency oriented; structured around organizational goals; performance metrics; technology-mediated & 
Defined by expertise/position & 
Strong results orientation \\
\midrule

\textbf{Survival Scenario} & 
CPS in time-critical, high-stakes situations requiring rapid coordination & 
High-stakes decision making; predefined protocols; rapid information processing; minimal deliberation & 
Highly structured, authority-based & 
Immediate problem resolution \\
\midrule

\textbf{Game-Based Collaboration} & 
CPS in game environments requiring coordinated problem solving & 
Engagement driven; artificial constraints; explicit rules; balanced challenges; feedback mechanisms & 
Player-chosen & 
Both process enjoyment and winning \\
\bottomrule
\end{tabular}
}
\caption{Collaborative problem-solving contexts and characteristics}
\label{tab:cps-contexts}
\end{table*}

\begin{table*}[t!]
\centering
\resizebox{1.0\linewidth}{!}
{
\begin{tabular}{p{23.0cm}}
\hline
\\
Analyze dialogue utterances in collaborative contexts. Classify the utterance into one of three Regulated Behavior types: Self-Regulated Learning (SRL), Co-Regulated Learning (CRL), or Socially Shared Regulated Learning (SSRL).\\
\\
=====\\
Regulated Behavior Types:\\
Type 1. Task Understanding\\
- Definition: Identifying and defining the problem; Understanding what information is already available; Investigating root causes\\
- Examples:\\
  * SRL: "I need to figure out what this problem is asking for."\\
  * CRL: "Do you agree that we need to understand what's being asked?"\\
  * SSRL: "I think we should also check what resources are already available to us."\\
\\
Type 2. Planning\\
- Definition: Establishing team structure and distributing tasks strategically\\
- Examples:\\
  * SRL: "I'm going to look at the data first."\\
  * CRL: "Let's split up the work. Maria, could you handle the research part?"\\
  * SSRL: "We could create a shared document to track progress. Does that work?"\\
\\
Type 3. Goal Setting\\
- Definition: Setting objectives for problem solving; Determining criteria for selecting the best solution\\
- Examples:\\
  * SRL: "I will find a solution that reduces costs by at least 15\%."\\
  * CRL: "Our objective should be to reduce costs. Does everyone agree we should aim for a 15\% reduction?"\\
  * SSRL: "I think our primary goal should be cost reduction, but we should also consider customer satisfaction."\\
\\
Type 4. Monitoring\\
- Definition: Tracking progress, summarizing completed work, and identifying remaining tasks\\
- Examples:\\
  * SRL: "I've completed the first part of the analysis."\\
  * CRL: "Have you finished the analysis section, Sam?"\\
  * SSRL: "I noticed we still need to address the risk section. Should we reassign that or adjust our timeline?"\\
\\
Type 5. Strategy Use\\
- Definition: Proposing potential solutions; Evaluating options and selecting the best solutions\\
- Examples:\\
  * SRL: "I think approach A is the best option because it addresses our key constraints."\\
  * CRL: "We could try solution A or B. Which do you all prefer? I think A might work better for our needs."\\
  * SSRL: "A is more cost-effective but B offers better scalability. What if we combine elements from both? Maria, you mentioned a hybrid approach earlier - could we integrate that with these options?"\\
\\
Type 6. Task Execution\\
- Definition: Implementing selected solutions\\
- Examples:\\
  * SRL: "I calculated the data based on our plan."\\
  * CRL: "Can you prepare the slides for reporting our findings?"\\
  * SSRL: "How about we create slides on Google Drive? We might need to adjust our approach based on what we discover during implementation."\\
\\
Type 7. Reflection\\
- Definition: Reflecting and making adjustments for continuous improvement\\
- Examples:\\
  * SRL: "I think I can recreate the scatter plot to visualize the result."\\
  * CRL: "Our solution didn't work as expected. Should we try the alternative method we discussed earlier?"\\
  * SSRL: "I think our solution didn't fully address the problem. Probably because we overlooked the seasonal variations. I remember Sarah mentioned this point before, right?"\\
\\
Type 8. None-of-Above\\
- Definition: The utterance does not present any meta-cognitive behavior.\\
=====\\
\\
Give me the type numbers and full type names. Provide your classification with a very concise explanation.\\
\\
\hline
\end{tabular}
}
\caption{One instruction example for collaborative dialogue analysis - Classify the utterance into one of three Regulation types.}
\label{appendix-inst-example}
\end{table*}

\end{document}